\documentclass[11pt,a4paper]{article}
\usepackage[hyperref]{acl2018}
\usepackage{graphicx}
\usepackage{times}
\usepackage{latexsym}
\usepackage{natbib}

\usepackage{url}

\aclfinalcopy 

\title{Machine Translation with Cross-lingual Word Embeddings}

\author{Marco Berlot \\
  {\tt mb2589@cornell.edu} \\\And
  Evan Kaplan \\
  {\tt emk269@cornell.edu} \\}

\date{}

\begin{document}
\maketitle
\begin{abstract}
  Learning word embeddings using distributional information is a task that has been studied by many researchers, and a lot of studies are reported in the literature. On the contrary, less studies were done for the case of multiple languages. The idea is to focus on a single representation for a pair of languages such that semantically similar words are closer to one another in the induced representation irrespective of the language. In this way, when data are missing for a particular language, classifiers from another language can be used.
\end{abstract}

\section{Introduction}

Methods for machine translations have been studied for years, and at the same time algorithms to generate word embeddings are becoming more and more accurate. Still, there is a lot of research aiming at unifying word embeddings across multiple languages. In this experience we try a technique for machine translation that relates word embeddings between two different languages. Based on the literature we found that it is possible to infer missing
dictionary  entries  using  distributed  representations
of words and phrases. One way of doing it is to create a linear mapping between the two vector spaces of two different languages. In order to achieve this, we first built two dictionaries of the two different languages. Next, we learned a function that projects the first vector space to the second one. In this way, we are able to translate every word belonging to the first language into the second one. Once we obtain the translated word embedding, we  output  the  most  similar
word vector as the translation. The word embeddings were learnt using the Skip Gram method proposed by (Mikolov et al., 2013a). An example of how the method would work is reported in figure 1 and figure 2. After creating the word embeddings from the two dictionaries, we plotted the numbers in the two graphs using PCA. Figure 3 reports the results after creating a  linear mapping between the embeddings from the two languages. You can see how similar words are closer together.

\begin{figure}[h]
\centering
\includegraphics[width=3in]{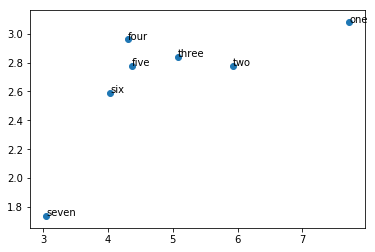}
\caption{PCA visualization of numbers in English}
\label{fig:fig1}
\end{figure}

\begin{figure}[h]
\centering
\includegraphics[width=3in]{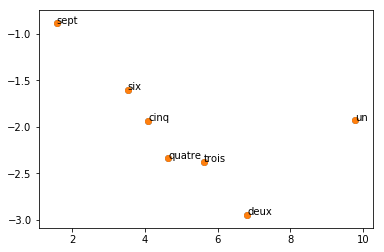}
\caption{PCA visualization of numbers in French}
\label{fig:fig2}
\end{figure}

\begin{figure}[h]
\centering
\includegraphics[width=3in]{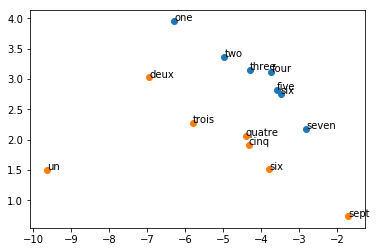}
\caption{PCA visualization of numbers in French and English after the linear mapping}
\label{fig:fig3}
\end{figure}

\section{Related Work}
In recent years, various models for learning cross-lingual representations have been proposed. Two main broad categories with some related papers are identified here:\\
\begin{itemize}
  \item \textbf{Monolingual mapping: }In this approach, models are trained using word embeddings from a monolingual corpora. Then, an objective function is used to minimize a linear mapping that enable them to map unknown words from the source language to the target language. \footnote{\\Linear projection (Mikolov et al., 2013),Projection via CCA (Faruqui and Dyer, 2014).}
  \item \textbf{Pseudo-cross-lingual: } In this case we create a pseudo-cross-lingual corpus by mixing contexts of different languages. We then train an off-the-shelf word embedding model on the created corpus. Ideally the cross-lingual contexts allow the learned representations to capture cross-lingual relations. \footnote{Mapping of translations to same representation (Xiao and Guo, 2014),Random translation replacement (Gouws and Sogaard, 2015)}

\end{itemize}
\section{Dataset}
In the literature, two main types of datasets are used for machine translation: Word-aligned data and Sentence-aligned data. The first one is basically a dictionary between the two languages, where there is a direct relation between same words in different languages. The second one has the relation between corresponding sentences in the two languages. We decided to start with the sentence aligned corpus, since it was more interesting to infer dependency from contexts among words.
For our experiment we decided to use the Europarl dataset, using the data from the WMT11 .The Europarl parallel corpus is extracted from the proceedings of the European Parliament. It includes versions in 21 European languages: Romanic (French, Italian, Spanish, Portuguese, Romanian), Germanic (English, Dutch, German, Danish, Swedish), Slavik (Bulgarian, Czech, Polish, Slovak, Slovene), Finni-Ugric (Finnish, Hungarian, Estonian), Baltic (Latvian, Lithuanian), and Greek. For this experience, we used the English-French parallel corpus, which contains 2,007,723 sentences and the English-Italian corpus, that contains 1,909,115 sentences.

\section{Linear Mapping}

This represents the linear function that maps word embeddings from one language to another. Given a word x in one language, and the respective word in the other language z, the following equation reports the objective function we want to minimize:
\begin{equation}\label{o1}
\sum\limits_{i=1}^n  |Wx_i-z_i|^2
\end{equation}

Were W represents the final matrix that will contain the values for the mapping.

\section{Normalization}
Since the word embeddings in a single language are based on the cosine similarity, we realized that through the objective function reported in equation 1 we were losing the property of this similarity. As a result we wanted the dot products to be preserved after the mapping. In order to do that we normalized the vectors $x_i$ and $z_i$.The new objective function looks like:
\begin{equation}\label{ob}
\sum\limits_{i=1}^n  |W\frac{x_i}{|x_i|}-\frac{z_i}{|z_i|}|^2
\end{equation}

In order to preserve the dot products after the linear mapping we also had to constrain W to be an orthogonal matrix. In order to orthogonalize the matrix, it's required to solve the following optimization problem:
\begin{equation}
\min |W-W'| \textrm{   s.t.   } W'{{W'}^T}=I
\end{equation}
One  can  show  that  this  problem  can  be  solved  by
taking  the  singular  value  decomposition  (SVD)  of
W and replacing the singular values to ones. This approximation would only work when the dimensions of the source vector and the target vector are the same, which is the set up we are working with. So finally, the method that we tried for this experience uses normalized vectors, and is reduced by using the cosine similarity function.
\section{Objective Function}
Taking equation \ref{ob} we expanded it getting the following equation:
\begin{equation}
\sum\limits_{i=1}^n  |\frac{W{x_i}}{|Wx_i|}|^2+|\frac{z_i}{|z_i|}|^2-2\frac{Wx_i}{|Wx_i|}^T\frac{z_i}{|z_i|}
\end{equation}
Which is really easy to show that through some simplifications will result into:\\
\begin{equation}\label{o2}
\textrm{argmax}\sum\limits_{i=1}^n \cos({Wx_i,z_i})
\end{equation}
Where cos represents the cosine similarity between the two embeddings. 
\section{Setup Description}
For this experience we tried monolingual mapping using the Europarl Dataset and the μtopia parallel corpus \footnote{Microblogs as Parallel Corpora, Wang Ling, Guang Xiang, Chris Dyer, Alan Black and Isabel Trancoso, ACL 2013}.
Concerning the preprocessing, we tokenized the text into single words, and every number was substituted with a NUM symbol. In addition, all the special characters were removed. To obtain the dictionaries, we used the words from the English corpora and translated them into the target languages using the Google translate API. In this way we built two dictionaries with corresponding words in the two languages, extracted from the same parallel corpora.

\section{Methods}
\subsection{Skipgram}
It was recently shown that the distributed representations of words capture surprisingly many linguistic regularities, and that there are many types of
similarities among words that can be expressed as
linear translations (Mikolov et al., 2013c).
In the Skip-gram model, the
training objective is to learn word vector representations  that  are  good  at  predicting  its  context  in  the
same  sentence  (Mikolov  et  al.,  2013a). The objective function that skip gram tries to minimize is the following:\\ $$\sum\limits_{i=1}^N \sum\limits_{y=-k}^{+k}  \log P (w_{i+j} | w_i)$$
Where N represents the total number of words in a sentence, P the probability of a word at the position i+j to belong to the sentence, with respect to the word i. Ideally, by using this approach we will be able to provide non-trivial translations that will be related to the context of a word.
\subsection{Minimizing the loss function}
 In order to minimize the loss function we decided to setup a neural network and reduce the objective by using Stochastic Gradient Descent. Figure 4 represents the architecture of our base model. We repeated the same process both for objective functions expressed in equation \ref{o1} and \ref{o2}. We started by getting an English word from the English corpus and we got the corresponding target word by using Google Translate. If this  word was contained in the target corpus, this pair of words was used to train the model; otherwise, it was ignored. We then get the two embeddings and get the result by multiplying the first one by the matrix W and then substracting the second embedding. For this experiment, we used embeddings with a dimension of 100. For this reason W is a matrix of size 100*100. 

\begin{figure}[h]
\centering
\includegraphics[width=4in]{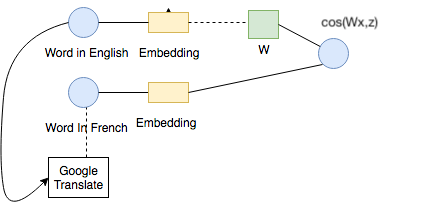}
\caption{Model architecture used to reduce the objective function}
\label{fig:fig3}
\end{figure}
\section{Results}
\subsection{Accuracy}
The process for checking the accuracy of the model consisted of first taking a random subset of the corpus that was not used for training. This subset consists of slightly more common words because the more common words should have more accurate word embeddings. Infrequent words (like websites and serial numbers) would not necessarily have accurate word embeddings that were generated through the word2vec model. The English word embedding was multiplied by the transformation matrix to create the predicted translated word embedding. The cosine similarity was then generated with each translated word embedding stored in the corpus. The 20 words with the highest cosine similarity were then outputted. Then the English word was translated through Google Translate and compared to the 20 outputted translated words. It was considered a match if the translated word was found in this list of 20 similar words. The reason why the translated word was not just compared to the most similar embedding was because each word could have multiple semantic meanings or have different synonyms in the other language. After running this accuracy formulation, results for different languages and objective functions are reported in Table \ref{tab:t1}. In the table you can see the different results we obtained across different languages. You can notice that we increased the number of languages from the first review of our paper. In addition, the table shows the comparison between our baseline method (Least Squares) and our final one (Normalized vectors, with cosine similarity), which shows a slight improvement with respect to the first one.\\
\begin{table}
\begin{tabular}{|l|c|r|}
	\hline
	Language & Least Squares & Normalized Vectors \\
	\hline
    French	& 47.1\% & 49.3\% \\
	\hline
    Italian	& 36.4\% & 39.2\% \\
	\hline
    Spanish	& 45\% & 47.24\% \\
	\hline
    Chinese	& 1.1\% & 1.4\% \\
	\hline

\end{tabular}
\caption{\label{tab:t1}Comparison of results using the baseline method (Least Squares) and the Normalized Vectors}

\end{table}
\\

\section{Error Analysis}
\subsection{Chinese Translations}
It's easy to notice how the performance with the Chinese translations was much lower with respect to the other languages. One of the reason is that even the tokenization of the chinese language is not trivial. While using a standard library, we noticed that some words were tokenized ambiguously. Another problem was getting the specific translation from Google translate. A lot of words might have very similar meanings, and Google Translate is not as accurate as it is with the other languages we have worked with. One reason is that Chinese is structured in a very different way compared to the other western languages, which share common roots. For example, a simple word like "Yes", does not have a direct translation in Chinese.
\subsection{Google Translate}
We noticed that the model was performing worse than expected. For this reason we studied what was the main source of error. It was interesting that the model was able to always predict very close French words in French that had the same meaning as the English ones. The problem was that they did not completely match the ones obtained from Google translate. In fact, it resulted that often, the Google translated word was not the most accurate one. Table 2 reports two of these examples. The first one is represented by the word help. All the three translations are pretty accurate, especially the first one, which is the right translation for the verb "to help". The problem is the translation provided by Google, which literally means "Help Me".
The second example reports the word "fire". In French there are different words to express the concept of a fire, the concept of an apartment "on fire" or the verb "fire a gun". All the translations provided by the model represent the different meanings that the English word has.

This problem was alleviated by translating the 20 outputted French words back into English and comparing those with the original English word. By doing this process, the accuracy of the system increased to 47\%. We expected the results of the cosine similarity objective function to produce a higher accuracy, but in reality, we a achieved slightly worse results.
\\

\begin{tabular}{|l|c|r|}
	\hline
	English Word & Model Translations & Google\\
	\hline
    help	& aider & Aidez-moi\\
	\hline
    	 & motiver &  \\
    \hline
    	 & encourager & \\
	\hline
    fire	& incendie & feu\\
	\hline
    	 & pistolet &  \\
    \hline
    	 & fusil & \\
	\hline

\end{tabular}
\\

\section{The Code}
The code can be found at \url{https://github.com/MarcoBerlot/Languages_for_Machine_Translation}. The Predictive model file contains all the implementations, from the feature engineering to the training of the model.

\section{References}
Linear projection (Mikolov et al., 2013) \\\\	Lexicon
Projection via CCA (Faruqui and Dyer, 2014)\\\\
Normalisation and orthogonal transformation (Xing et al., 2015)\\
Alignment-based projection (Guo et al., 2015)\\

\end{document}